\documentclass[letterpaper, 10 pt, conference]{ieeeconf}  % Comment this line out if you need a4paper

\IEEEoverridecommandlockouts                              % This command is only needed if 
\title{\LARGE \bf
Imagination at Inference: Synthesizing In-Hand Views for \\
Robust Visuomotor Policy Inference
}
\author{Haoran Ding$^{1}$, Anqing Duan$^{1}$, Zezhou Sun$^{1}$, Dezhen Song$^{1}$, Yoshihiko Nakamura$^{1}$ % <-this % stops a space
\thanks{
$^{1}$Haoran Ding, Anqing Duan, Zezhou Sun, Dezhen Song and Yoshihiko Nakamura are with the Department of Robotics, Mohamed bin Zayed University of Artificial Intelligence (MBZUAI), Abu Dhabi, UAE. {\tt\small \{haoran.ding, anqing.duan, zezhou.sun, dezhen.song, yoshihiko.nakamura\}@mbzuai.ac.ae}} 
}

\usepackage{xcolor}
\usepackage{cite}
\usepackage{graphicx}
\usepackage{amsmath,amsfonts}
\usepackage{booktabs}
\usepackage{textcomp}

\begin{document}

\maketitle
\thispagestyle{empty}
\pagestyle{empty}

%%%%%%%%%%%%%%%%%%%%%%%%%%%%%%%%%%%%%%%%%%%%%%%%%%%%%%%%%%%%%%%%%%%%%%%%%%%%%%%%
\begin{abstract}
Visual observations from different viewpoints can significantly influence the performance of visuomotor policies in robotic manipulation. Among these, egocentric (in-hand) views often provide crucial information for precise control. However, in some applications, equipping robots with dedicated in-hand cameras may pose challenges due to hardware constraints, system complexity, and cost. 
% However, in some practical scenarios, equipping robots with dedicated in-hand cameras may be impractical due to hardware constraints, system complexity, and cost. 
In this work, we propose to endow robots with imaginative perception — enabling them to \textquotesingle imagine\textquotesingle\ in-hand observations from agent views at inference time. We achieve this via novel view synthesis (NVS), leveraging a fine-tuned diffusion model conditioned on the relative pose between the agent and in-hand views cameras. Specifically, we apply LoRA-based fine-tuning to adapt a pretrained NVS model (ZeroNVS) to the robotic manipulation domain. We evaluate our approach on both simulation benchmarks (RoboMimic and MimicGen) and real-world experiments using a Unitree Z1 robotic arm for a strawberry picking task. Results show that synthesized in-hand views significantly enhance policy inference, effectively recovering the performance drop caused by the absence of real in-hand cameras. Our method offers a scalable and hardware-light solution for deploying robust visuomotor policies, highlighting the potential of imaginative visual reasoning in embodied agents.
\end{abstract}

%%%%%%%%%%%%%%%%%%%%%%%%%%%%%%%%%%%%%%%%%%%%%%%%%%%%%%%%%%%%%%%%%%%%%%%%%%%%%%%%

\section{Introduction}
Robotic manipulation requires robust perception and decision-making. Visuomotor policies, which map visual observations to control actions, have shown promise in enabling end-to-end control for tasks such as object picking, placement, and assembly~\cite{e2evisuomotor, kalashnikov2018scalable, diffpol}. These policies typically rely on RGB inputs captured from onboard, external, or combined camera views~\cite{chi2024universal, octo_2023, li2024visionlanguage} to condition their control decisions. The effectiveness of such policies strongly depends on the quality and informativeness of the visual input. 
Specifically, the camera's field of view, perspective, and occlusion characteristics significantly influence the agent's perceptual capabilities and, consequently, its performance~\cite{jangir2022look, akinola2020learning}.
% In particular, the field of view, perspective, and occlusion characteristics of the camera play a critical role in what the agent can perceive and thus how well it can act~\cite{jangir2022look, akinola2020learning}.

\begin{figure}[t]
    \centering
    \includegraphics[width=\linewidth]{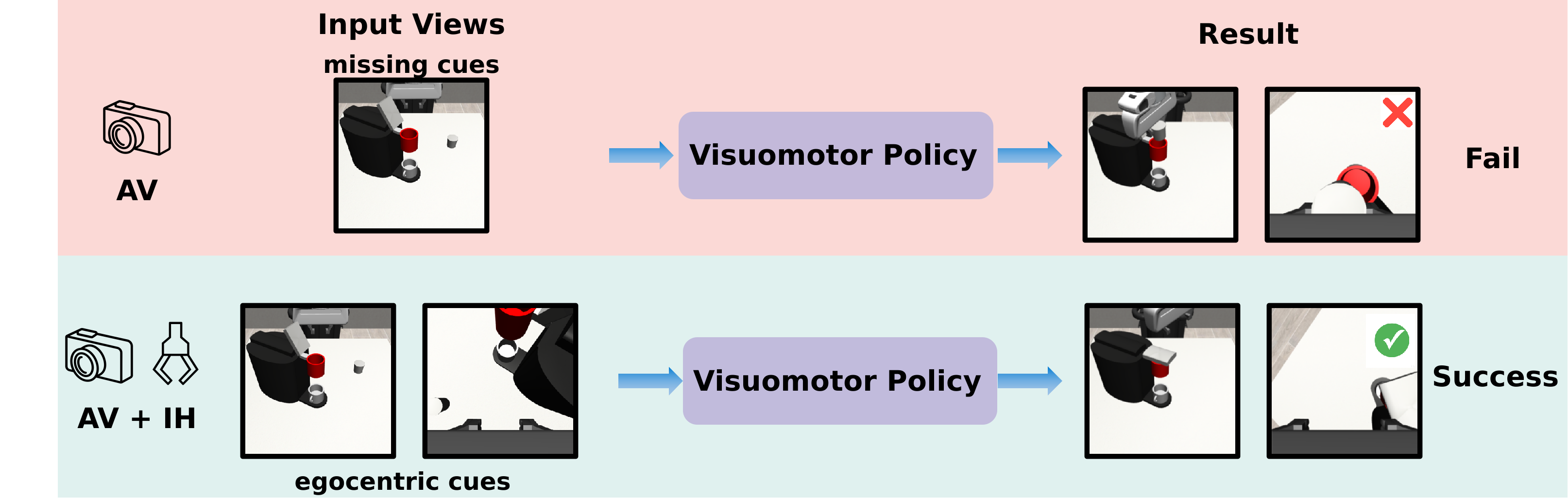}
    \caption{Comparison of policy executions on the MimicGen Coffee task. Top: policy with agent view only. Bottom: policy augmented with in-hand observations. Without in-hand input, the policy in the first row fails to place the object correctly; with in-hand observations, the policy in the second row achieves successful placement.}
    \label{fig:policy compare}
\end{figure}

Recent work has demonstrated that incorporating multiple viewpoints, such as combining an external (agent) view with a wrist-mounted (in-hand) camera, significantly improves manipulation performance by providing richer, complementary information~\cite{jangir2022look, acar2023visual, chuang2024active}. For example, the use of eye-in-hand views reduces ambiguity during close-proximity tasks like grasping or insertion by offering detailed, task-relevant information~\cite{akinola2020learning, cheng2018reinforcement}, as shown in Figure~\ref{fig:policy compare}. However, deploying physical in-hand cameras on real robots poses several challenges. Mounting a camera on the end-effector can introduce mechanical interference or restrict the robot's range of motion~\cite{seo2023multi}. Additional hardware cost and cabling can increase system complexity, and occlusions from the gripper or object can limit the usability of the view~\cite{cheng2018reinforcement}. 
% Accurate hand–eye calibration is typically required to align the in-hand camera frame with the robot, which is both tedious and susceptible to error.

Inspired by recent advances in generative modeling --- particularly single-image, camera pose-conditioned novel view synthesis (NVS) techniques such as Zero123~\cite{liu2023zero}, ZeroNVS~\cite{zeronvs}, and VISTA~\cite{tian2024viewinvariant} --- we explore a new opportunity for enabling in-hand perception in robotics. These approaches can synthesize novel views of a scene from a single input image and a specified camera pose, producing 3D-consistent renderings without the need for dense multi-view supervision or explicit geometry. While such techniques have primarily been used for offline augmentation or learning viewpoint-invariant policies\cite{tian2024viewinvariant, zhou2023nerf}, we repurpose them for online use during inference. This enables the robot to generate in-hand views on the fly, effectively expanding its visual input space without adding any additional physical hardware.

In this paper, we operationalize this idea by integrating camera pose-conditioned novel view synthesis (NVS) into visuomotor policy inference. Our approach trains a diffusion-based NVS model to synthesize in-hand views from an external RGB observation and a known relative transformation between the two viewpoints. The synthesized in-hand view is then provided as an additional input to the policy at test time, enabling dual-view inference without the need for a physical wrist-mounted camera. This framework is modular and can be seamlessly integrated into existing visual policy pipelines, and while our experiments focus on imitation learning, the proposed method is, in principle, compatible with other learning paradigms. 

Compared to prior work in the robotics area that uses NVS as offline data augmentation~\cite{tian2024viewinvariant, zhou2023nerf, dcoda, shi2025nvspolicy}, our method performs online view synthesis at runtime, providing live visual information that fills the missing egocentric perspective. We evaluate our system on diverse manipulation tasks in both simulation (RoboMimic~\cite{robomimic2021} and MimicGen~\cite{mimicgen}) and real robot settings (Unitree Z1 robot arm), showing that policies equipped with synthesized in-hand views significantly outperform those using external views only, and closely match policies that access real wrist-mounted camera input.

To summarize, our contributions are:
\begin{itemize}
    \item[$\bullet$] We establish the critical role of in-hand viewpoints in visuomotor policy performance and motivate replacing physical wrist cameras with synthesized views as a conceptual alternative.
    % We highlight the significant contribution of in-hand viewpoints to visuomotor policy performance and propose a scalable alternative to physical camera deployment by synthesizing such views using generative models. 
    \item[$\bullet$] We adapt a pre-trained pose-conditioned diffusion model (ZeroNVS) to robotic manipulation domains via parameter-efficient LoRA fine-tuning, enabling the synthesis of in-hand views from agent views. 
    \item[$\bullet$] We demonstrate how synthesized in-hand views can be seamlessly integrated into real-time policy inference, improving policy success rates without modifying or retraining the policy. 
\end{itemize}

\section{related work}
Effective visuomotor policy learning requires the ability to perceive and act robustly across varying camera viewpoints. Prior work has explored this challenge from multiple angles, including training with diverse camera inputs, learning viewpoint-invariant visual representations, and generating novel observations through view synthesis. In this section, we review relevant efforts in multi-view policy learning and novel view synthesis in robotics.
\subsection{Visuomotor Policy Learning with Multi-View Observation}
Multi-view observation has proven to be an effective strategy for enhancing the robustness and generalization of visuomotor policies. By capturing the scene from diverse viewpoints, these approaches mitigate occlusions, resolve spatial ambiguities, and enable the learning of viewpoint-invariant visual representations. Sermanet et al.~\cite{sermanet2018time} propose Time-Contrastive Networks (TCN), a self-supervised method that learns viewpoint-invariant features from multi-view videos, enabling third-person imitation by aligning demonstrations to an egocentric frame. Akinola et al.~\cite{akinola2020learning} leverage multiple uncalibrated cameras in reinforcement learning, combined with sensor dropout, to train policies that are robust to missing or noisy views.

Subsequent studies provide empirical evidence on the benefits of multi-view setups. Ablett et al.~\cite{ablett2021seeing} demonstrate that policies trained with multiple camera views generalize better to unseen viewpoints compared to single-view baselines. Jangir et al.~\cite{jangir2022look} introduce a Transformer-based framework that fuses third-person and wrist-mounted views, outperforming single-view policies on complex manipulation tasks. More recently, Hsu et al.~\cite{hsu2022vision} systematically study the impact of different viewpoints, finding that egocentric in-hand views significantly boost both learning speed and generalization, especially when coupled with information bottlenecks to reduce overfitting to third-person views. 

To alleviate the hardware burden of multi-camera systems, Acar et al.~\cite{acar2023visual} propose knowledge distillation from multi-view teacher policies to single-view students, achieving strong performance with reduced sensory inputs. In a complementary direction, Seo et al.~\cite{seo2023multi} train multi-view masked world models that learn view-invariant features, enabling robust sim-to-real transfer.

Our work builds upon these insights by emphasizing the importance of in-hand egocentric observations. Distinct from prior approaches that require physical in-hand cameras during deployment, we instead synthesize such views from external observations at test time. This allows for robust policy deployment without introducing additional hardware complexity.

\subsection{Novel View Synthesis for Robotics Manipulation}
Novel view synthesis (NVS) has emerged as a powerful tool in computer vision, enabling image generation from unseen viewpoints based on a limited set of observations. In the robotics domain, NVS techniques have recently been adopted to improve visual perception, pose estimation, and even policy learning. Among the earliest applications, Lin et al.~\cite{yen2021inerf} propose iNeRF, which inverts pre-trained neural radiance fields to estimate camera poses by minimizing photometric reconstruction errors, enabling 6-DoF localization without explicit feature matching.

NeRF-based representations have also been used to facilitate the manipulation of visually challenging objects. Ichnowski et al.~\cite{ichnowski2021dex} introduce Dex-NeRF, which leverages NeRF’s implicit geometry to estimate graspable surfaces on transparent objects. Evo-NeRF~\cite{kerr2022evo} extends this idea by incrementally updating the NeRF representation as objects are removed, enabling efficient sequential grasping in cluttered scenes. These works highlight the use of NVS to overcome depth-sensing limitations in real-world robotic manipulation.

\begin{figure*}[t]
    \centering
    \includegraphics[width=\linewidth]{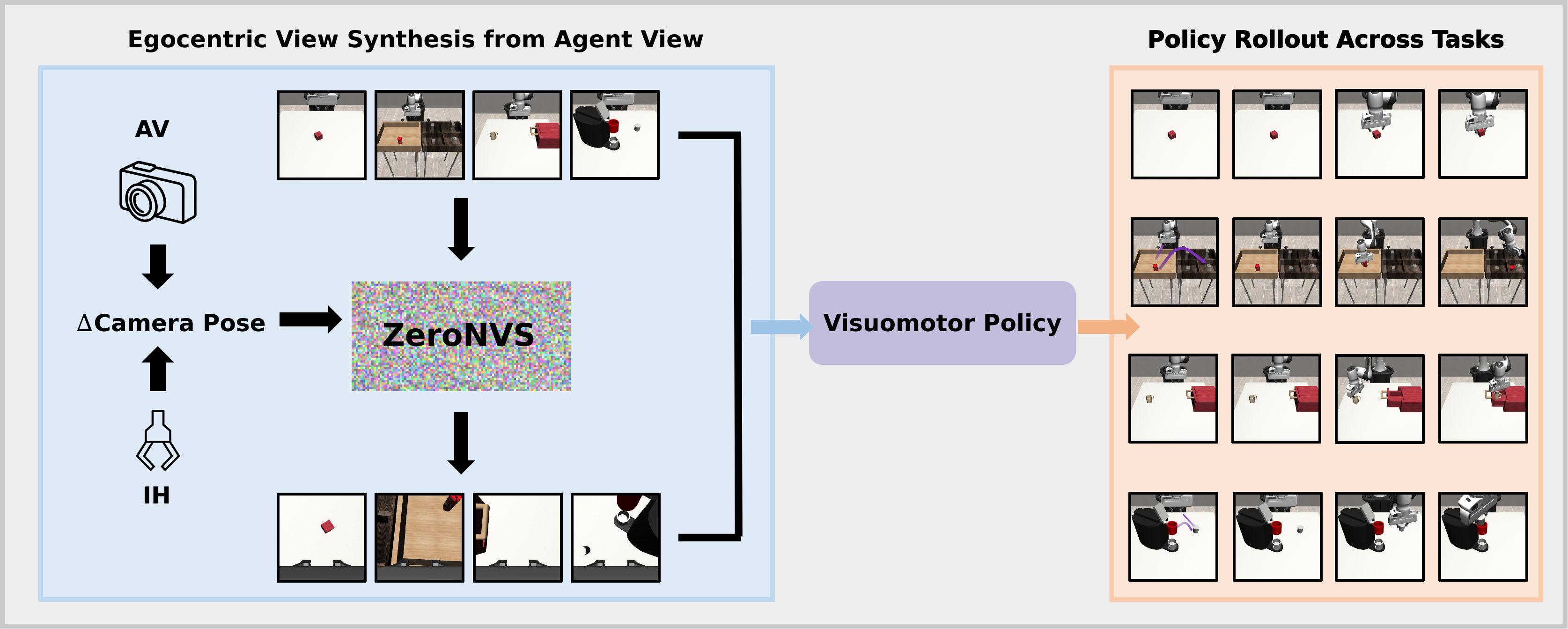}
    \caption{Method overview. Our approach generates egocentric in-hand views using ZeroNVS, conditioned on external agent-view observations and relative camera pose transformations. These synthesized views complement external inputs and are used for policy deployment across various manipulation tasks.}
    \label{fig:method overview}
\end{figure*}

Beyond perception, NVS has also been employed to augment visuomotor policy learning. Zhou et al.~\cite{zhou2023nerf} propose SPARTN, which synthesizes off-demonstration viewpoints using NeRF and pairs them with corrective actions to improve the robustness of imitation policies. In a similar spirit, Li et al.~\cite{shi2025nvspolicy} introduce NVSPolicy, which actively selects and synthesizes auxiliary views during execution using a generative model, enabling the policy to “imagine” additional perspectives and better adapt to scene variation.

These methods demonstrate the potential of integrating NVS into robotic pipelines—not just for visual understanding but also for enhancing robustness and adaptability of downstream control. Our work builds on this trend by leveraging diffusion-based NVS model to synthesize egocentric observations during deployment, effectively bypassing the need for in-hand cameras while maintaining strong visuomotor performance.

\section{methodology}
Our method is motivated by the observation that visuomotor policies benefit significantly from incorporating both agent and in-hand camera views\footnote{Throughout this paper, we refer to the fixed third-person camera as the \textit{agent view (AV)}, and the wrist-mounted egocentric camera as the \textit{in-hand view (IH)}.}. However, equipping a physical robot with a wrist-mounted in-hand camera can be costly, physically intrusive, or infeasible in certain deployment scenarios~\cite{cheng2018reinforcement}. To address these constraints, we employ a camera pose-conditioned novel view synthesis (NVS) model to generate in-hand observations from agent views and known camera transformations. This enables us to retain the benefits of in-hand perception without requiring additional sensing hardware at test time. 

In the following subsections, we formalize the problem and the train–test view mismatch in Sec.~\ref{subsec: problem def}. Sec.~\ref{subsec: nvs diff} presents the pose-conditioned diffusion model used to synthesize in-hand observations from external views and known extrinsics. Sec.~\ref{subsec: lora} details LoRA fine-tuning on task-specific data. Sec.~\ref{subsec: policy} describes how the synthesized views are integrated into the visuomotor policy for inference.
\subsection{Problem Definition}
\label{subsec: problem def}
To enable synthesis of in-hand observations from agent views, we begin by formalizing the problem setting and clarifying the data available during training and deployment.

We consider a robot manipulation scenario in which each observation consists of an RGB image from an agent-view camera $I_{AV} \in \mathbb{R}^{H \times W \times 3}$, and, during training, an aligned image from an in-hand camera $I_{IH} \in \mathbb{R}^{H \times W \times 3}$. The spatial relationship between the two views is described by the relative transformation $T_{AV \to IH}\in SE(3)$, computed from calibrated camera poses.

The training dataset is composed of triplets: $D_{train} = \{(I_{AV}^{(i)}, I_{IH}^{(i)}, T_{AV \to IH}^{(i)})\}_{i=1}^N$, which we use to train a pose-conditioned generative model $G$ to synthesize in-hand views from agent views: $\hat{I}_{IH} = G(I_{AV}, T_{AV\to IH}; \theta)$,
% \begin{equation}
%     \hat{I}_{IH} = G(I_{AV}, T_{AV\to IH}; \theta),
% \end{equation}
where $\hat{I}_{IH}$ is synthesized in-hand view observation, $\theta$ is the parameter of $G$. 

At policy test time, in-hand views are unavailable. Instead, the synthesized view $\hat{I}_{IH}$ is used alongside the external observation as input to a visuomotor policy $\pi$, which outputs the control action: $a=\pi(I_{AV}, \hat{I}_{IH})$. This formulation allows policies to exploit in-hand information at deployment without relying on additional sensing hardware.

\subsection{Novel View Synthesis via Pose-Conditioned Diffusion}
\label{subsec: nvs diff}
To generate in-hand observations from agent views and known relative poses between two cameras, we adopt a diffusion-based novel view synthesis model based on ZeroNVS~\cite{zeronvs}. ZeroNVS builds upon Zero123~\cite{liu2023zero}, which itself extends the Stable Diffusion --- a latent diffusion framework --- with pose conditioning, and performs single-view, pose-conditioned image synthesis without requiring 3D supervision or multi-view inputs. This makes it particularly suitable for robotic settings, where only monocular RGB images and pose information are available during deployment.

Given an agent-view image $I_{AV}$ and a relative pose $T_{AV \to IH}$, the model first encodes the image into a latent feature space using a variational autoencoder. The $6$-DoF pose is embedded through a shallow MLP and broadcast spatially to form a pose conditioning tensor. These two inputs are jointly processed in a U-Net denoiser via cross-attention layers at each diffusion timestep, guiding the generative process toward the desired viewpoint. After iterative denoising, the synthesized latent code is decoded into the target in-hand image $I_{IH}$. 

This architecture enables ZeroNVS to implicitly learn 3D scene structure from 2D pose-labeled data. Compared to NeRF-based methods~\cite{nerf}, it is faster at inference and significantly easier to integrate into robotics pipelines, requiring only a single image and pose as input while producing consistent and realistic viewpoint-conditioned outputs.

\subsection{LoRA Fine-tuning}
\label{subsec: lora}
To adapt the pre-trained ZeroNVS model to the domain of robotic manipulation, we fine-tune the latent diffusion model ZeroNVS using Low-Rank Adaption (LoRA)~\cite{lora}, implemented via the PEFT library~\cite{xu2023parameter}. Instead of updating all model weights, LoRA introduces trainable low-rank matrices into the attention modules of the U-Net denoiser. This enables efficient domain adaptation while minimizing memory and computational overhead. Specifically, we apply LoRA to the query, key, value, and output projection layers in both cross-attention and self-attention blocks, using a rank of $128$, a scaling factor of $256$, and a dropout of $0.1$. All other model parameters remain frozen during training.

The model is optimized using a denoising loss defined as the mean squared error between predicted and ground-truth noise vectors across sampled diffusion steps. Training data consists of paired RGB images from two viewpoints, along with the relative camera transformation in $SE(3)$, which is used for pose conditioning.

We adopt standard optimization settings for training, including gradient accumulation and checkpoint selection based on validation performance. This tuning strategy enables effective domain adaptation without full model retraining, while remaining efficient in both compute and memory.

\subsection{Integration into Policy}
\label{subsec: policy}
The synthesized in-hand observations are used to enable dual-view policy inference in settings where a physical wrist-mounted camera is not available. During training, the policy receives both the agent-view camera image $I_{AV}$ and the real in-hand observation $I_{IH}$, together with task-specific observations. At test time, the in-hand camera is removed, and the policy instead receives a synthesized observation $\hat{I}_{IH}$ generated by the view synthesis model based on the external view and relative pose.

The policy architecture is designed to support dual-view visual input by processing the agent-view and in-hand observations through separate image encoders. The resulting feature embeddings are fused and passed to an action prediction module, which may be instantiated as a diffusion-based decoder, a behavior cloning head, or other task-appropriate structures. This modular design enables compatibility with a range of policy learning algorithms. In this paper, we mainly focus on imitation learning policy, but this method could also be used for reinforcement learning.

\section{experiment}
To evaluate the effectiveness of using synthesized in-hand views for visuomotor policy deployment, we conduct a comprehensive set of experiments in both simulated and real-world environments. Specifically, we benchmark our approach on six manipulation tasks: three from RoboMimic~\cite{robomimic2021} (\textsc{lift}, \textsc{can}, \textsc{square}) and three from Mimicgen~\cite{mimicgen} (\textsc{stack}, \textsc{coffee}, \textsc{mug cleanup}). We further validate its practical applicability through a real-world strawberry picking task using a Unitree Z1 robot arm. 

Across all settings, we compare three policy input configurations: (1) AV: agent view only, (2) AV + GT-IH: agent view and ground-truth in-hand view (available during training and test-time ablation), and (3) AV + Syn-IH: agent view and synthesized in-hand view. This evaluation setup allows us to assess whether synthesized observations can effectively recover the policy performance gains typically associated with in-hand camera input. We further include ablation studies to analyze the impact of model design choices and augmentation strategies.
\subsection{Experimental Setup}
\begin{figure*}[t]
    \centering
    \includegraphics[width=0.32\linewidth]{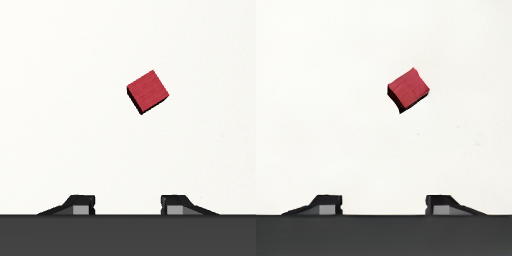}
    \includegraphics[width=0.32\linewidth]{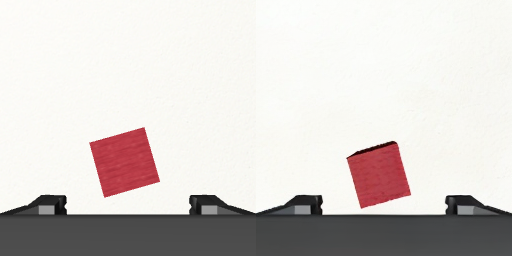}
    \includegraphics[width=0.32\linewidth]{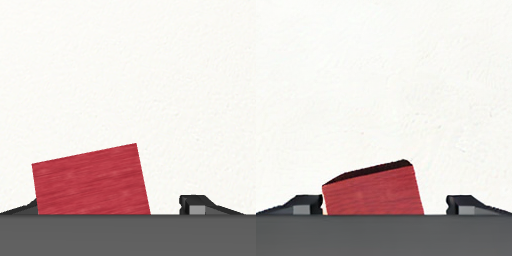} \\
    \vspace{2mm}
    \includegraphics[width=0.32\linewidth]{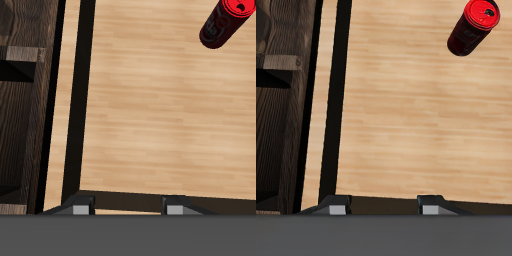}
    \includegraphics[width=0.32\linewidth]{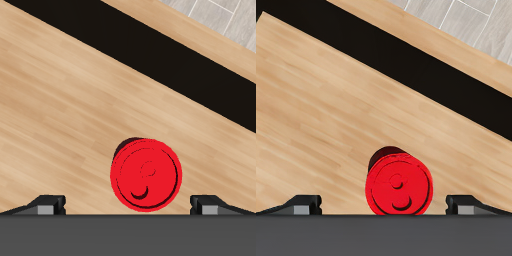}
    \includegraphics[width=0.32\linewidth]{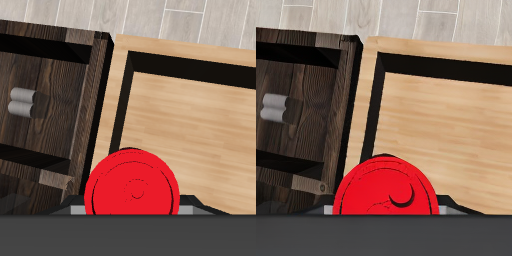} \\
    \vspace{2mm}
    \includegraphics[width=0.32\linewidth]{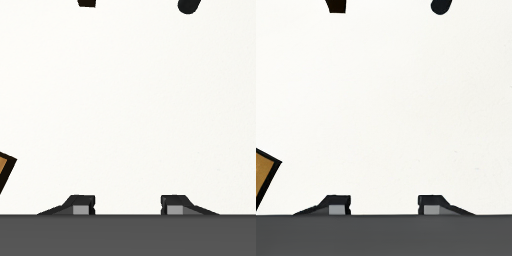}
    \includegraphics[width=0.32\linewidth]{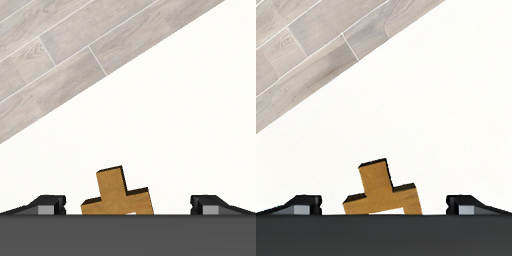}
    \includegraphics[width=0.32\linewidth]{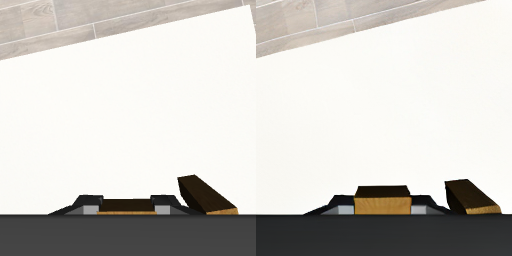}\\
    \vspace{2mm}
    \includegraphics[width=0.32\linewidth]{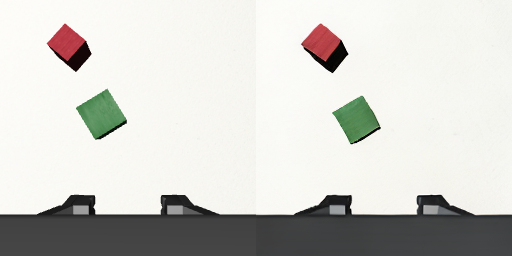}
    \includegraphics[width=0.32\linewidth]{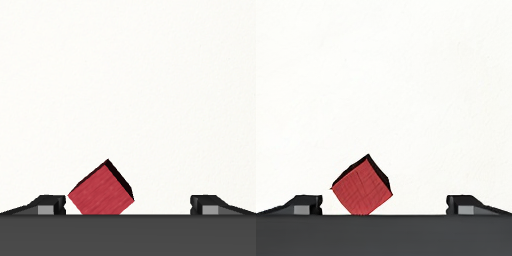}
    \includegraphics[width=0.32\linewidth]{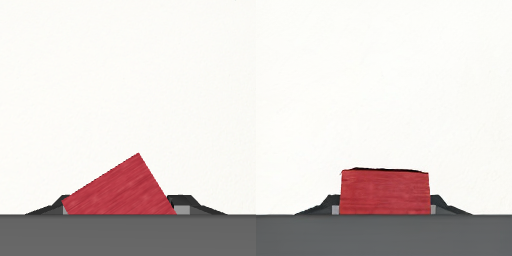} \\
    \vspace{2mm}
    \includegraphics[width=0.32\linewidth]{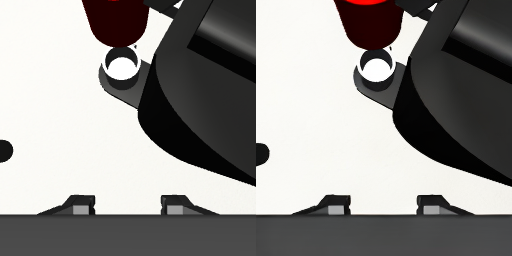}
    \includegraphics[width=0.32\linewidth]{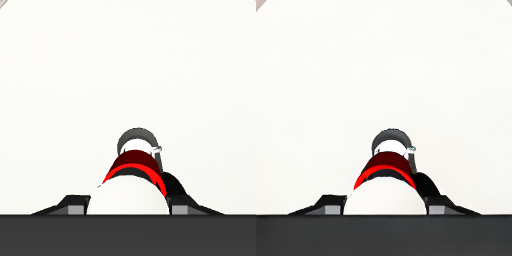}
    \includegraphics[width=0.32\linewidth]{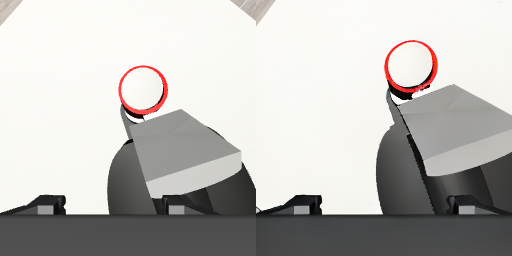} \\
    \vspace{2mm}
    \includegraphics[width=0.32\linewidth]{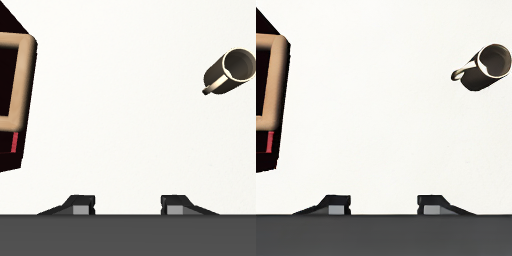}
    \includegraphics[width=0.32\linewidth]{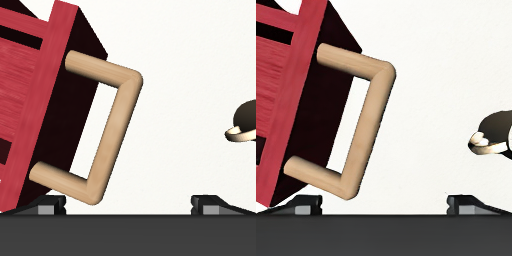}
    \includegraphics[width=0.32\linewidth]{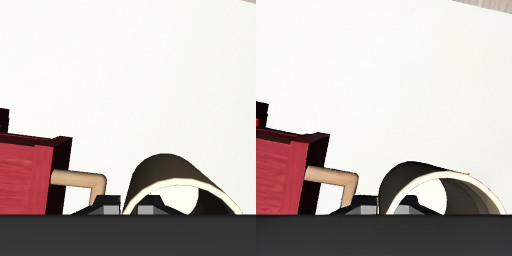}
    \caption{Qualitative comparison of synthesized in-hand views across six manipulation tasks. Each pair of images shows the ground-truth in-hand view (left) and the synthesized result (right), generated from a known agent-view observation and relative camera pose. The tasks include \textsc{Lift}, \textsc{Can}, and \textsc{Square} from the RoboMimic benchmark (top three rows), and \textsc{Stack}, \textsc{Coffee}, and \textsc{Mug Cleanup} from the MimicGen benchmark (bottom three rows).}
    \label{fig:syn_vs_gt}
\end{figure*}

To implement our proposed framework, we separately configure the policy learning pipeline and the view synthesis module. In our experiments, we adopt Flow Matching Policy (FMP)~\cite{braun2024riemannian, funk2024actionflow, fang2025flow} as the primary visuomotor policy. Similar to Diffusion Policy~\cite{diffpol}, FMP is based on a generative modeling framework --- flow matching~\cite{lipman2022flow} --- for trajectory prediction, but offers improved training stability and faster inference, making it well-suited for both simulation and real-world deployment~\cite{ding2024fast}. We train two series of policies: one conditioned only on external camera inputs (external-only), and one conditioned on both external and in-hand views (dual-view). Each policy receives inputs from either an agent-view camera or both agent and in-hand views, depending on the experimental configuration. 

In our implementation, the visual encoder for each view is a ResNet-18 backbone, and the image features are concatenated with robot proprioception signals (end-effector position, orientation, and gripper state). The resulting conditioning vector is passed to a UNet-based architecture, where different tasks use varying UNet sizes according to scene complexity.

To provide synthesized in-hand views during policy deployment, we fine-tune a ZeroNVS-based latent diffusion model using Low-Rank Adaptation (LoRA). We inject LoRA modules into the self- and cross-attention layers of the UNet denoiser, specifically targeting the 'to\_q', 'to\_k', 'to\_v', and 'to\_out.0' projections. The LoRA hyperparameters are set to a rank of 128 and a scaling factor of 256. Fine-tuning is conducted with MSE denoising loss over 4 NVIDIA A100 GPUs, using RGB image pairs with relative camera poses from RoboMimic and MimicGen datasets.

% During training, all policies are provided with real in-hand and external views. 
At policy test time, the synthesized in-hand image generated from the agent's view and the known camera transformation is used as a substitute. We standardize image resolution to $84\times 84$ (Robomimic \& Mimicgen) or $180 \times 320$ (real robot experiment)  for policy training, and apply data augmentations such as random cropping (with crop percentage $90 \%$). Hyperparameters such as learning rate, batch size, and training horizon are specified per task.

\subsection{Simulation Results: Robomimic and Mimicgen}
\begin{figure*}[t]
    \centering
    \includegraphics[width=0.98\linewidth]{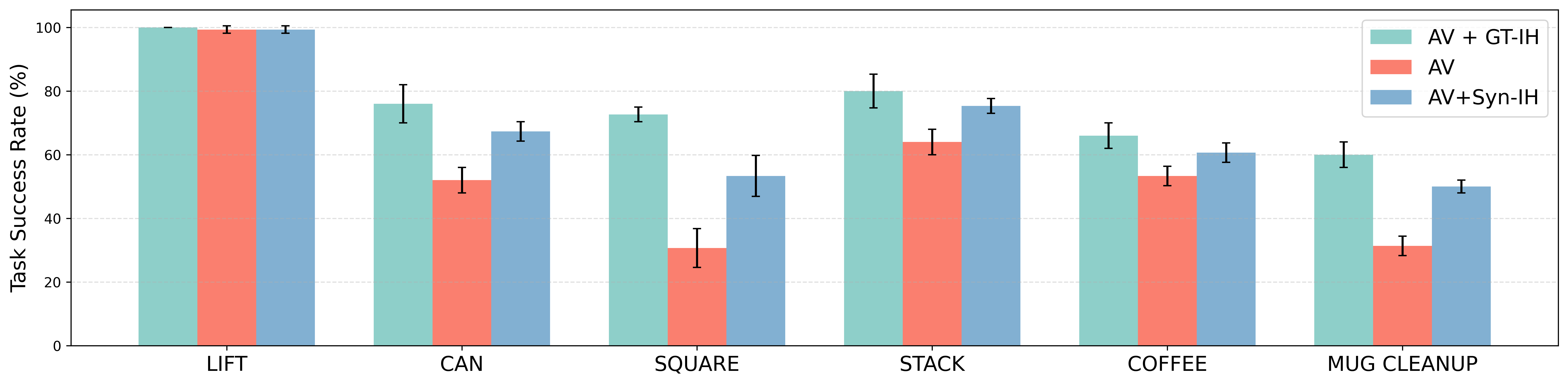}
    \caption{Success Rates by Tasks under different observation input: AV + GT-IH: agent view and ground-truth in-hand view; AV: agent view; AV + Syn-IH: agent view and synthesized in-hand view}
    \label{fig:simulator result}
\end{figure*}
We report task success under three settings: AV (agent view only), AV+GT-IH (agent view with ground-truth in-hand), and AV+Syn-IH (agent view with synthesized in-hand). For AV+Syn-IH, we adopt a two-stage evaluation to quantify the incremental benefit of synthesized views: we first roll out the policy with AV only; on failures, we re-run the same trials with Syn-IH enabled and count the additional recoveries. This metric isolates the contribution of synthesized in-hand observations while keeping the architecture and rollout protocol fixed. We report the mean success over 50 evaluation seeds per configuration. For fairness, the same 50 seeds (initial states and simulator randomness) are reused across AV, AV+GT-IH, and AV+Syn-IH. For each configuration, we perform three evaluation repeats per seed and report the overall mean and standard deviation; the grouped bar charts display the mean with error bars denoting ±1 std.

\textbf{Task Description}
\begin{itemize}
    \item \textsc{Lift}: grasp a single object and lift it to a target height.
    \item \textsc{Can}:  reach to and manipulate a small cylindrical can to a designated target.
    \item \textsc{Square}: align and insert a square peg/object into a square receptacle.
    \item \textsc{Stack}: grasp a block and place it stably on top of another block.
    \item \textsc{Coffee}: insert a coffee capsule into the coffee machine and close the lid.
    \item \textsc{Mug Cleanup}: open a drawer, pick up the mug, place it into the drawer, and close the drawer.
\end{itemize}
All tasks use an observation horizon of 2 frames. The prediction horizon is 64 for RoboMimic and 32 for MimicGen. We train Flow Matching Policies (FMP) with batch size 256 throughout. RoboMimic policies are trained for 200 epochs. For MimicGen, Stack uses 200 epochs, while Coffee and Mug Cleanup use 1000 epochs. For MimicGen, we intentionally cap the data at 200 demonstrations per task (rather than all 1000), consistent with prior findings that 200 demonstrations are sufficient to attain strong imitation performance on these tasks~\cite{funk2024actionflow, wang2025equivariant}. The policy backbone, encoders, and optimization settings are kept fixed across settings to ensure comparability.

Across tasks, the AV+GT-IH configuration establishes an upper bound on policy performance, while AV+Syn-IH reflects how much of this performance gap can be recovered without access to a physical wrist-mounted camera. As illustrated in Figure~\ref{fig:simulator result}, synthesized in-hand views consistently improve success rates compared to the AV-only baseline across all six tasks. In particular, tasks involving occlusion and fine-grained spatial reasoning—such as \textsc{Square}, \textsc{Mug Cleanup}, and \textsc{Can}—benefit most from egocentric input, with AV+Syn-IH closing a substantial portion of the gap toward AV+GT-IH. For simpler tasks such as \textsc{Lift}, all configurations achieve near-saturated performance, suggesting that the contribution of synthesized views is task-dependent.

While AV+Syn-IH improves task success across the board, inference-time view synthesis incurs non-negligible compute overhead. In our implementation, synthesizing a single in-hand view with ZeroNVS takes approximately 16.5 s per frame on an NVIDIA RTX 4060 Ti. This added latency may limit closed-loop responsiveness in the deployment and underscores the importance of efficiency when integrating generative models into control pipelines.
\begin{figure}[t]
    \centering
    \includegraphics[width=1.0\linewidth]{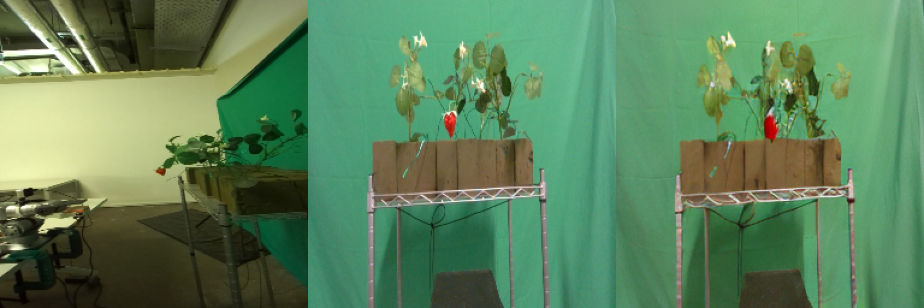}
    \caption{Real-world Unitree Z1 Strawberry Picking experiment: agent view, ground-truth in-hand view, and ZeroNVS-synthesized in-hand view.}
    \label{fig:unitree}
\end{figure}

\begin{table}[t]
\centering
\caption{Unitree Strawberry Picking Success Rate}
\begin{tabular}{lccc}
\toprule
 & \textbf{Agent} & \textbf{Agent + GT-Inhand} & \textbf{Agent + Syn-Inhand}\\
\midrule
Success Rate & $3 / 10$ & $8/10$ & $6/10$\\
\bottomrule
\end{tabular}
\label{table: unitree}
\end{table}
\subsection{Real-World Evaluation: Strawberry Picking}
A Unitree Z1 manipulator with a parallel gripper operates in a benchtop workspace containing plastic strawberries that are magnetically attached to a slender branch, producing a reproducible detachment (“pluck”) when sufficient normal force is applied, as shown in Figure~\ref{fig:unitree}. Sensing comprises an external ZED~2i camera providing the agent view (AV) and a wrist-mounted Intel RealSense D405 providing the in-hand view (IH). The cameras are time-synchronized and extrinsically calibrated; all images used for policy inference are resized to $180 \times 320$. During finetuning of the ZeroNVS model, images are instead resized to $256\times 256$ to match ZeroNVS training resolution. Each trial begins from a fixed initial pose with the branch positioned within reach, while the target strawberry’s pose is varied across trials. The objective is to approach the target, establish a stable grasp, and detach it from the branch, followed by a short lift to verify a secure pick. We evaluate three configurations: AV, AV+GT-IH, and AV+Syn-IH, as in previous simulation experiments. A trial is counted as a success if the robot cleanly detaches the strawberry and holds it stably after lift; failures include missed grasps, slips after contact, or incomplete detachment.

For this experiment, we collected 10 demonstrations by teleoperation. Each demonstration logs synchronized image pairs from the in-hand camera RealSense D405 (native $480 \times 848$) and the agent camera ZED 2i (native $720 \times 1280$), together with robot joint state and griper state. We additionally calculate the calibrated extrinsic transform between the two cameras, $T_{AV \to IH} \in SE(3)$ (represented as a $4 \times 4$ matrix), by combining an offline hand-eye calibration and the logged gripper poses. Specifically, we first calibrate the wrist camera w.r.t. the gripper to obtain $T_{gripper \to IH}$, and measure the cross-camera alignment at the initial frame $T_{Agent \to IH}(t_0)$. Given the gripper pose $T_{W \to gripper}(t_0)$ recorded in the rosbag, we recover the fixed external camera pose:
{
\small
\begin{equation}
    T_{W \to AV} = T_{W \to gripper} (t_0) T_{gripper \to IH} [T_{AV \to IH}(t_0)]^{-1}.
\end{equation}}
$T_{W \to AV}$ and $T_{W \to gripper}$ denote the poses of agent- and in hand-view cameras in the world coordinate frame. Then for any time $t$,
{
\small
\begin{equation}
    T_{AV \to IH}(t) = T^{-1}_{W\to AV} T_{W \to gripper}(t) T_{gripper \to IH}.
\end{equation}
}
This yields a time-varying, calibrated relative pose between the external (agent) and in-hand views throughout the whole demonstration.
For policy training, both views are resized by bilinear interpolation to $180 \times 320$ and augmented with random cropping at a $0.9$ scale factor to improve robustness. The policies use an observation of $1$ frame and a prediction horizon of $32$ steps. The base network is a 3-layer UNet with down dimensions $(256, 512, 1024)$. 

At test time, we evaluate policy performance across 10 distinct strawberry positions within the workspace. The positions are selected to induce variation in depth, lateral offset, and orientation while remaining within the training distribution. To ensure fair comparison, we place the target strawberry in approximately the same pose across different policy configurations (AV vs AV+Syn-IH vs AV + GT-IH), using visual alignment with the branch as a reference. The surrounding foliage and physical background are kept unchanged from the demonstration recordings, and the experiment is conducted under fixed indoor lighting conditions to control for visual domain shift.

Results are summarized in Table~\ref{table: unitree}. Out of 10 trials, the AV-only policy succeeds in 3 cases. Augmenting the policy with synthesized in-hand views improves performance to 6/10, while the ground-truth IH configuration achieves 8/10. This confirms that synthesized in-hand views can enhance policy stability and robustness at inference time, despite not fully matching the upper bound provided by real wrist camera input. The improvement over the AV-only baseline demonstrates the practical value of view synthesis in real-world deployment.

% \subsection{Ablations}
\begin{figure}[t] 
    \centering
    \includegraphics[width=0.95\linewidth]{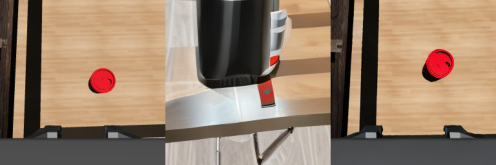} \\
    \vspace{2mm}
    \includegraphics[width=0.95\linewidth]{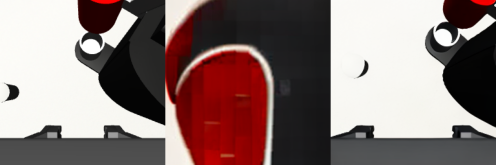} 
    \caption{ Qualitative comparison of synthesized in-hand views before and after LoRA finetuning of ZeroNVS. Top: \textsc{can} task. Bottom: \textsc{Coffee} task. From left to right: ground-truth in-hand view, pretrained ZeroNVS output, and finetuned output (ours). Finetuned models recover task-relevant geometry, while pretrained models fail to generate meaningful egocentric views, often omitting critical information present in the in-hand perspective.}
    \label{fig:finetune effect}
\end{figure}

% \textbf{Effect of ZeroNVS Lora Finetuning: }
\subsection{Effect of ZeroNVS Lora Finetuning}
We qualitatively evaluate the impact of finetuning the ZeroNVS model on task-specific data. Figure~\ref{fig:finetune effect} compares synthesized in-hand views from the pretrained model and our LoRA-finetuned variant, alongside ground-truth in-hand images. We visualize two representative tasks: \textsc{Can} from RoboMimic and \textsc{Coffee} from MimicGen. The pretrained model often fails to capture scene structure, producing blurred or severely distorted outputs. In contrast, the finetuned model yields sharp, geometrically consistent views that closely match the true egocentric observations. These improvements are especially pronounced near contact events and object boundaries, which are critical for precise manipulation.

% \begin{table}[t]
% \centering
% \caption{Impact of View Configurations on Diffusion Policy}
% \begin{tabular}{lccc}
% \toprule
% Tasks & \textbf{Agent} & \textbf{Agent + GT-Inhand} & \textbf{Agent + Syn-Inhand}\\
% \midrule
%  \textsc{can}&  & & \\
%  \textsc{Coffee} & & & \\
% \bottomrule
% \end{tabular}
% \label{table: diffpol}
% \end{table}

% \textbf{Effect on Diffusion Policy: }To assess whether the benefit of synthesized in-hand views (Syn-IH) extends beyond our primary backbone (FMP), we instantiate a second visuomotor policy using Diffusion Policy (DP). We keep the data budget, observation/prediction horizons, visual encoders but with longer training epoch, and train DP on two representative tasks: \textsc{Can} (RoboMimic) and \textsc{Coffee} (MimicGen). At test time, we evaluate the same configurations—AV, AV+GT-IH, and AV+Syn-IH.

% \textbf{Crop Percentage Ablation} \todo{maybe add the table of this ablation}

\section{conclusion}
In this paper, we identify the limitations of relying solely on external agent-view observations for visuomotor policy inference. Leveraging advances in novel view synthesis (NVS), we fine-tuned a diffusion-based ZeroNVS model on both simulated (RoboMimic, MimicGen) and real-world (Unitree Z1 strawberry picking) manipulation datasets. Our experiments show that incorporating synthesized in-hand views at inference time consistently improves policy robustness and task success, narrowing the gap toward policies equipped with physical wrist cameras. These results suggest that NVS can serve as a lightweight and scalable alternative to additional sensing hardware for vision-conditioned policies under partial observability.
% Motivated by advances in computer vision, we explore the use of novel view synthesis (NVS) to reconstruct missing egocentric (in-hand) perspectives from an external view and known camera geometry. To this end, we fine-tune a latent diffusion-based NVS model (ZeroNVS) on diverse manipulation datasets, including six simulated tasks from the RoboMimic and MimicGen benchmarks and a real-world strawberry-picking task using a Unitree Z1 robot. Across all settings, our experiments demonstrate that incorporating synthesized in-hand views at inference time significantly improves policy robustness and task success, narrowing the gap toward the performance of policies equipped with physical wrist cameras. These results suggest that NVS can serve as a lightweight and scalable alternative to instrumenting robots with additional sensing hardware, and open new directions for vision-conditioned policy deployment under partial observability.

However, our current implementation based on ZeroNVS introduces noticeable latency during inference, which can hinder smooth closed-loop execution. Reducing generation time, for example through model distillation~\cite{song2023consistency}, is an important direction for future work. Moreover, our pipeline still requires in-hand views during policy training and ZeroNVS fine-tuning; this dependency could be alleviated by emerging 3D-free NVS approaches that synthesize controllable camera viewpoints from a single image without multi-view supervision~\cite{kang20243d}.
\bibliographystyle{IEEEtran}
\bibliography{ref}

\end{document}